# Design of reliable technology valuation model with calibrated machine learning of patent indicators

*Seunghyun Lee[a], Janghyeok Yoon[a], Jaewoong Choi[b,*]*

[a]Department of Industrial Engineering, Konkuk University, Seoul 05029, Republic of Korea

[b]Computational Science Research Center, Korea Institute of Science and Technology, Seoul 06130, Republic of Korea

[*]Corresponding author: jwchoi95@kist.re.kr

**Abstract**

Machine learning (ML) has revolutionized the digital transformation of technology valuation by predicting the value of patents with high accuracy. However, the lack of validation regarding the reliability of these models hinders experts from fully trusting the confidence of model predictions. To address this issue, we propose an analytical framework for reliable technology valuation using calibrated ML models, which provide robust confidence levels in model predictions. We extract quantitative patent indicators that represent various technology characteristics as input data, using the patent maintenance period as a proxy for technology values. Multiple ML models are developed to capture the nonlinear relationship between patent indicators and technology value. The reliability and accuracy of these models are evaluated, presenting a Pareto-front map where the expected calibration error, Matthews correlation coefficient and F1-scores are compared. After identifying the best-performing model, we apply SHapley Additive exPlanation (SHAP) analysis to pinpoint the most significant input features by confidence bin. Through a case study, we confirmed that the proposed approach offers a practical guideline for developing reliable and accurate ML-based technology valuation models, with significant implications for both academia and industry.

# 1. Introduction

In the domain of technology valuation, machine learning (ML) is gaining prominence as a cutting-edge approach. ML models have been used to capture the nonlinear relationships between patent indicators, which represent technological characteristics, and the value of the technology (Kim et al., 2022; Ko et al., 2019). Specifically, most prior works use forward citations (Hong et al., 2022; Lee et al., 2018), technology transactions (Kim et al., 2023; Ko et al., 2019; Lee et al., 2023), or patent maintenance (Choi et al., 2020) as proxies for technology values, thereby evaluating the economic value or technological impact of patents. These approaches not only ensure reasonable performance but also accelerate the process of expert technology valuation, making it more efficient and objective. The adoption of ML models has thus enabled the digital transformation of technology valuation (Kim et al., 2022).

However, the reliability of these ML-based technology valuation models has not been thoroughly validated. These models tend to exhibit overconfidence, displaying high certainty even when their predictions are incorrect (Guo et al., 2017). In the context of technology valuation, where model and data uncertainties are significant (Choi et al., 2019), providing trustworthy information about the confidence levels of predictions is crucial. This issue can lead to substantial problems in practical applications, resulting in misguided decisions based on overly confident yet inaccurate assessments. Therefore, further research is necessary to address the reliability and confidence calibration of these models to ensure their practical utility and accuracy in real-world scenarios.

As a remedy, we propose a calibrated ML-based approach for reliable technology valuation based on the expected calibration error (ECE) score and the SHapely Additive exPlanations (SHAP) analysis. First, we employ patent maintenance information as a new proxy of valuable technologies, as long-term maintenance of patent rights is indicative of business value of individual patents and is domain-agnostic (Choi et al., 2020). Also, 50 quantitative patent indicators – such as specificity, scientific intensity or patent family – which are directly or potentially related to technology values, are utilized as inputs. Next, we develop various ML models to capture the nonlinear relationships between patent lifetimes and multiple patent indicators. Third, we evaluate the performance of ML models with quantitative measures such as ECE, and Matthews correlation coefficient (MCC), presenting a Pareto-front map to identify the most reliable and accurate model. Finally, for the best-performing model, we apply the SHAP analysis to the prediction results by confidence bin, thereby identifying the most significant input features on the prediction uncertainty.

We applied the proposed approach to 46,973 patents related to semiconductor technologies. The USPTO database was used for this research, as it provides a high-quality dataset with comprehensive bibliographic data, citations, and text information. Our experiments demonstrated that the proposed approach can identify the most reliable and accurate ML models to predict the value of technologies using early patent indicators. Additionally, our SHAP analysis revealed that in regions of low confidence, the competitiveness of relevant technological fields has a significant negative impact on technological values. Conversely, in regions of high confidence, the activity of relevant technological fields has a substantial positive impact. The results of our case study provide practical guidelines for obtaining reliable and accurate ML models for technology valuation. We anticipate that the proposed approach will be a valuable complementary tool to support experts' decision-making in technology valuation by providing reliable and interpretable results. This systematic process and scientific outcomes can facilitate expert-machine collaboration in real technology environments characterized by high uncertainty.

The remainder of this paper is organized as follows: Section 2 provides the background of our research. Section 3 details the proposed approach, which is illustrated with an empirical example in Section 4. Section 5 discusses the implementation and customization of the proposed approach, along with its implications for theory, practice, and policy. Finally, Section 6 concludes with the limitations of our study and suggests directions for future research.

## 2. Background

### 2.1 ML-based technology valuation approaches

The existing literature in technology innovation has introduced various patent indicators to represent the heterogeneous characteristics of valuable technologies. Patent-based approaches have capitalized on empirical findings that there is a significant difference of quantitative patent indicators between valuable technologies and less valuable ones. Compared to experts-based technology valuation methods, these approaches benefited the consistency, and objectivity of outcomes derived from a large volume of patents and the reproducibility and operational efficiency in terms of methods. Patent-based approaches to technology valuation used various proxies of technology value, mainly patent forward citation, technology transactions and patent lifetime (Choi et al., 2020; Kim et al., 2019; Lee et al., 2018).

First, patent citation count has been used to signify the ripple effect of a technology on subsequent advancements, representing the impact a technology has on future innovations. This proxy is based on the premise that technologies with significant economic values or innovative ideas are likely to have substantial impact on follow-up technologies. Accordingly, early identification of patents expected to receive significant citations in the future provides strategic insights for idea screening, emerging technology detection, and technological impact analysis. In this regard, many approaches to predict the number of forward citations over the particular period have been proposed. For instance, Lee et al. (2018) employed artificial neural networks to capture the nonlinear relationships between forward citation counts and various patent indicators, which can be assessed promptly upon the issuance of relevant patents. Similarly, Hong et al. (2022) have employed word2vec and convolutional neural network models on patent abstract text to extract nuanced technical meanings and proficiently model their relationship with patent citation counts. Second, technology transaction data has been used to evaluate the economic value of patents, as it provides insights into how patented technologies are being utilized, licensed, or transferred in the marketplace. By analyzing technology transaction records with patent indicators, it would be possible to understand the economic impact and potential value of newly issued patents. For instance, Ko et al. (2019) proposed a deep neural network model that discriminates valuable patents with technology transfer records from less valuable ones using quantitative patent indicators. Similarly, Lee et al. (2023) suggested how to identify valuable universities and research institutes-oriented technologies with deep neural networks, showing that the number of domestic priority patents is the most significant input feature for technology transaction. Recently, Kim et al. (2022) introduced a SHAP analysis to examine the importance of patent indicators on the economic value of technologies, thereby providing both high predictability and interpretability.

Unfortunately, these two proxies are domain-variant, meaning that different thresholds are required to determine the quantitative standard of valuable technologies. Instead, we employ patent maintenance information as a proxy of valuable technologies, as long-term maintenance of patent rights is indicative of business value of individual patents. Patent holders are more likely to keep their patents if the profits generated, either directly or indirectly, exceed the maintenance fees needed to uphold the exclusive rights to the patent (Hikkerova et al., 2014). In addition, patent lifetimes can be divided into several pre-defined groups — approximately 4, 8, 12 or 20 after patent registration — according to the United States Patent and Trademark Office (USPTO). Using such information, Choi et al. (2020) suggested to predict whether a patent will survive until its maximum expiration date with quantitative patent indicators, revealing that the neural networks model is the best-performing ML model.

## 2.2. Quantitative patent indicators representing technology characteristics

Previous studies have defined patent indicators representing various technological characteristics which are directly or potentially related to the value of technologies. In this study, 50 quantitative indicators are defined based on exhaustive literature survey, which can be divided into six categories: scope and coverage, priority, completeness, development effort and capabilities, technology environment and prior knowledge. We focused what patent indicators measure and when they can be extracted from the database, as some indicators such as number of backward citations are lagging indicators and cannot be used as input of predictive models.

### 2.2.1. Scope and coverage

For technological scope and coverage of patents, seven quantitative indicators can be employed; specificity of total technical scope, international scope of prior patents, scope of technological components, peripheral components, and core components, and average specificity of core components. First, the specificity of technical scope is measured with the number of words contained in the patent's full text, as the extent of detailing the invention can reflect the efforts, time and expense invested in the invention as well as the volume of information (Choi et al., 2020; Trappey et al., 2012). The number of different nations in the backward citation information of the patent is used to reflect the international scope of prior patents, as the origins of prior technologies that underpin an invention can potentially impact the patent value. Generally, independent claims of a patent describe the most fundamental knowledge of the invention, while dependent claims provide extended information of the cited independent claims (Reitzig, 2004; Tong and Frame, 1994). In this regard, it has been reported that the more claims the patent includes, the more likely the patent is to pass the examination process and further have an influential value or quality (Fischer and Leidinger, 2014; Lanjouw and Schankerman, 2004). Moreover, considering that the type and number of claims significantly impact the cost of patent registration, information regarding the claims is deemed crucial in assessing the patent's value (Jeong et al., 2016; Lanjouw and Schankerman, 2001). Accordingly, the number of dependent, independent and total claims in the patent is calculated to reflect the scope of core, peripheral and total technological components. Similarly, the average numbers of words contained in independent claims in the patent are calculated to capture the average specificity of core technological components, assuming that the specificity of independent claims are potentially related to the legal scope or novelty of the invention. Lastly, the number of different IPCs assigned to the patent is used to consider the technical coverage in the system, as it has been reported that this measure can help to identify patent portfolios of assignees and further patent values (Chen and Chen, 2011; Lee et al., 2009).

### 2.2.2. Priority

The priority of a patent is crucial in assessing its value as it protects the initial filing date of an invention. Asserting this priority validates the early-stage innovation and uniqueness of the technology, thereby enhancing the patent's validity (Choi et al., 2020; Su et al., 2011). It underscores the importance of the patent's inception date in assessing its overall worth. In this regard, the number of priority patents of the patent and different nations where the patent has its priorities have been used to quantify the priority intensity and its international range.

### 2.2.3. Completeness

To consider the completeness of a patent as potential indicators of patent values, five different measures can be used; dependency of prior knowledge (domestic/foreign), grant time lag and specificity of technical summary. The number of prior patents that the patent cites at the time of issuance date is used as a potential indicator, despite the controversial relations between the dependency of prior knowledge and the patent values (Allison and Lemley, 1998; Lanjouw and Schankerman, 2001). Further, this measure is divided according to whether the prior patents are domestic or not, by calculating the number of US and non-US patents that the patent cites at the time of issuance date. The grant time lag of the patent, i.e., the interval between its filing date and issuance date representing the patent examination period, is used as a potential measure, as a long-term examination may indicate a great deal of effort of stakeholder to improve its patent (Hikkerova et al., 2014). Lastly, the specificity of technical summary is quantified based on the number of words contained in the patent's abstract, as the abstract includes the main contents of the invention and the extent of details can determine the scope or values of patents.

**2.2.4. Development effort and capabilities**

As the development effort and capabilities may have positive relationships on patent values, seven metrics can be employed as inputs. First, the number of assignees involved with the patent is counted to measure the size of contributors, as the events of patent maintenance are affected by the capabilities of assignees (Lai and Che, 2009). Similarly, we used the number of non-US assignees involved with the patent to quantify the contributions of foreign assignees. To reflect the international distribution of contributors, the number of different nations of assignees involved with the patent is calculated, as there could be the different tendency of patent maintenance or values by nationality. In terms of inventors, it has been reported that patents created by multiple inventors are likely to be more significant than those by a single inventor (Ernst, 2003; Ma and Lee, 2008). Therefore, the number of inventors, non-US inventors, and number of different nations of inventors are used to quantify the efforts of inventors. As a potential measure, the average number of overdue maintenance fees of assignees can be used to reflect the tendency of assignees in patenting activities.

**2.2.5. Technology environment**

As the external technological environment may have an effect on the decision making related to patent maintenance and R&D strategies (Choi et al., 2023; Trappey et al., 2012), environmental indicators have been used to capture such changes in the relevant technology fields for each patent. First, the average number of patents issued yearly in the relevant technology fields is used to capture how active the technology fields were at the issuance timing. Similarly, the average number of cumulative patents issued in the relevant technology fields is employed to identify the size of accumulated knowledge (Lai and Che, 2009). Also, the average number of applicants issuing patents by issuance year in the relevant year is calculated to reflect the competitiveness of technology fields (Fabry et al., 2006). Technology fields of the patent in the system are calculated by counting the frequency of IPC codes at the section level. Lastly, growth speed indicator is measured based on the median gap of the filing date between prior patents and the patent, known as technology cycle time (Bierly and Chakrabarti, 1996; Kayal and Waters, 1999), assuming that a fast-technological change of relevant technology fields can affect the decision-making of patent maintenance.

**2.2.6. Prior knowledge**

Regarding prior knowledge-related metrics, 10 metrics can be employed. First, the number of non-patent citations is employed to quantify scientific knowledge related to the invention, assuming that more scientific basis related to the patent may bring out more innovative and influential technologies (Cozzens et al., 2010; Rotolo et al., 2015; Trajtenberg, 1990). Next, the average number of patents issued by assignees and inventors are used to quantify prior experience of stakeholders, which can be important to patent value and maintenance. Similarly, core area know-how and peripheral area know-how are defined based on the prior patenting activities of assignees in the same technology field or other fields (Harhoff et al., 1999). This measure, so-called know-how, represents the level of accumulated knowledge of assignees, which can be interpreted as technological and commercial interest in a technology field of interest (Chung et al., 2021; Meyer, 2006). Specifically, we measured the number of patents issued by assignees in a technology field of interest or other fields as core area or peripheral know-how indicators (Seol et al., 2023). For the semantic similarity with prior patents, text embedding technique was applied to patent titles with sentence-level transformer models (Reimers and Gurevych, 2019), thereby calculating the similarity between the patent and its prior patents in the dense embedding space. Inclusiveness of technology fields are calculated based on the ratio of the IPC codes of the patent to those of prior patents, where IPC codes are analyzed at the subclass level (Lee et al., 2018). Technology fields of prior patents in the system are calculated for each section, based on the frequency of IPC codes. Technology breadth is defined based on the originality index (Bessen, 2008). Given a patent i, its prior patents P and a set of their IPCs N:

$$\text{Technology breadth}(i) = 1 - \sum_{n \in N} \frac{P(n)^2}{|P|}$$

where *P(n)* indicates the number of prior patents whose IPCs include *n*. Lastly, the dependency on homogeneous technologies is calculated by counting the number of prior patents in the same semiconductor technology field.

# 3. Methodology

In Fig. 1, the overall procedure of the proposed approach is illustrated, which consists of four discrete steps: (1) extracting patent indicators and technology value proxy, (2) estimating the technological values of patents, (3) screening the most reliable and accurate model, and finally (4) interpreting the model's predictions.

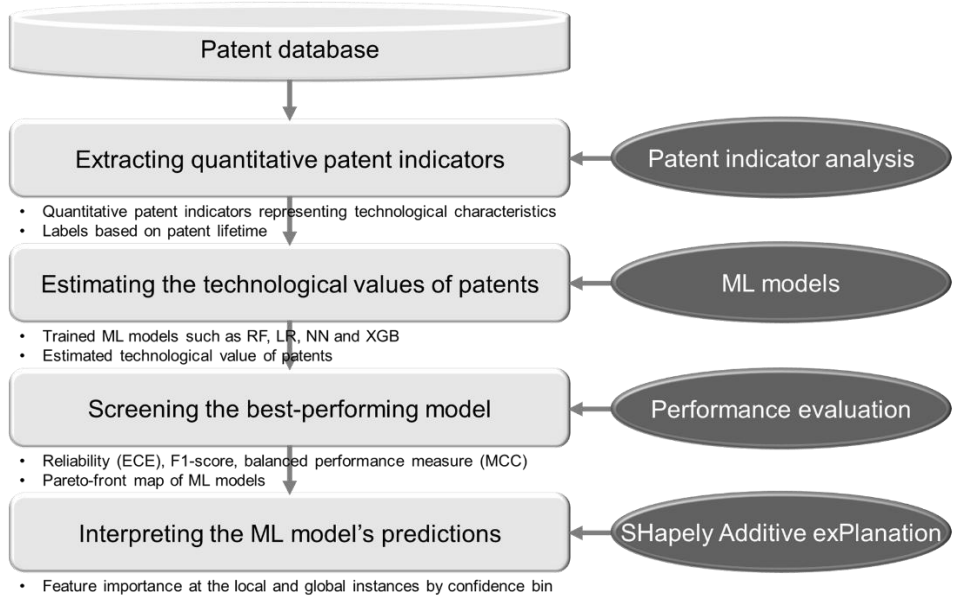

**Fig. 1.** Overall process of the proposed approach

## 3.1. Extracting patent indicators and technology value proxy

ML-based technology valuation models employ quantitative patent indicators, which have positive relationships with technology values. It is noteworthy that patent indicators can be determined at the time of patent valuation. Therefore, lagging indicators such as forward citation counts or technology transactions cannot be used as inputs for the valuation process. We selected 50 quantitative indicators directly or potentially related to technological values through an exhaustive literature survey (Table 2). These metrics can be divided into six categories with different perspectives; (1) scope and coverage, (2) priority, (3) completeness, (4) development effort and capabilities, (5) technology environment, and (6) prior knowledge. Among the indicators, there are 34 quantitative metrics, while the remaining 16 comprise technology field-related metrics are computed for each of the 8 sections of IPC (A-H).

**Table 1.** Patent indicators representing technological characteristics

| Category | Patent indicators | Operational description | Index |
|---|---|---|---|
| Scope and coverage | Specificity of total technical scope | Number of words contained in the patent's full text | SC_1 |
| | International scope of prior patents | Number of different nations in the backward citation information of the patent | SC_2 |
| | Scope of technological components | Number of total claims in the patent | SC_3 |
| | Scope of peripheral technological components | Number of dependent claims in the patent | SC_4 |
| | Scope of core technological components | Number of independent claims in the patent | SC_5 |
| | Average specificity of core technological components | Average numbers of words contained in independent claims in the patent | SC_6 |
| | Technical coverage in the system | Number of different IPCs assigned to the patent | SC_7 |
| Priority | Priority intensity | Number of priority patents of the patent | PR_1 |
| | International priority range | Number of different nations where the patent has its priorities | PR_2 |
| Completeness | Dependency of prior knowledge | Number of prior patents that the patent cites at the time of issuance date | CP_1 |
| | Dependency of domestic prior knowledge | Number of US patents that the patent cites at the time of issuance date | CP_2 |
| | Dependency of foreign prior knowledge | Number of non-US patents that the patent cites at the time of issuance date | CP_3 |
| | Grant time lag | Patent examination period | CP_4 |
| | Specificity of technical summary | Number of words contained in the patent's abstract | CP_5 |
| Development effort and capabilities | Size of contributors | Number of assignees involved with the patent | DEC_1 |
| | Contributions of foreign assignees | Number of non-US assignees involved with the patent | DEC_2 |
| | International distribution of contributors | Number of different nations of assignees involved with the patent | DEC_3 |
| | Efforts of inventors | Number of inventors involved with the patent | DEC_4 |
| | Efforts of foreign inventors | Number of non-US inventors involved with the patent | DEC_5 |
| | International cooperation degree | Number of different nations of inventors involved with the patent | DEC_6 |
| | Average number of surcharges by assignee | Average number of overdue maintenance fees of assignees | DEC_7 |
| Technology | Activity of technology fields | Average number of patents issued yearly in the relevant IPCs | TE_1 |
| environment | Size of technology fields | Average number of cumulative patents issued in the relevant IPCs | TE_2 |
| | Competitiveness of technology fields | Average number of applicants issuing patents by issuance year in the relevant IPCs | TE_3 |
| | Technology fields in the system | Frequency of sections A to H of the patent | TE_4 |
| | Growth speed | Median gap of the filing date between prior patents and the patent, as known as technology cycle time | TE_5 |
| Prior knowledge | Scientific knowledge | Number of non-patent citations | PK_1 |
| | Prior experience of assignees | Average number of patents issued by assignees | PK_2 |
| | Prior knowledge of inventors | Average number of patents issued by inventors | PK_3 |
| | Core area know-how | Number of patents in a technology field of interest issued by assignees | PK_4 |
| | Peripheral area know-how | Number of patents in other technology fields issued by assignees | PK_5 |
| | Semantic similarity with prior patents | Average text similarity with prior patents and the patent | PK_6 |
| | Inclusiveness of technology fields | The ratio of the IPC of the patent to the IPC of prior patents (based on subclass) | PK_7 |
| | Technology fields of prior patents in the system | Frequency of sections A to H of prior patents | PK_8 |

| | | |
|---|---|---|
| Technological breadth | Breadth of technological fields on which the patent is based | PK_9 |
| Dependency on homogeneous technologies | Number of backward citations in the semiconductor technology field | PK_10 |

## 3.2. Valuating the technological values of patents

In this step, various ML models can be used to predict the technological values of patents and we suggest to test the following representative models, i.e., logistic regression (LR), random forest (RF), standard neural networks (NN), and eXtreme Gradient Boosting (XGB) model (Refer to A1 for details). To derive reliable and accurate ML models, we apply several methods to perform a more effective training process. First, an under-sampling method is employed in this study. One of the aspects that negatively affect the ML model training process is related to class imbalance, where examples of training sets belonging to one class are much more numerous than examples from other classes. In the real world, the ratio of valuable and non-valuable technologies is not always the same. Under-sampling, which deals with class imbalance, is an efficient method for model training that uses only a subset of the majority class or removes data lying on fuzzy boundaries between classes. Next, k-fold cross-validation is also used. Cross-validation is one of the most widely used data resampling methods to assess a predictive model's generalization ability, tune model parameters, and prevent overfitting in the training process. In k-fold cross-validation, the available dataset is divided into k subsets of equal size. One subset is used for validation and the rest are used for training. This process is repeated until each subset is used as a validation set. The average performance of all the validation sets is given as the cross-validation performance for the predictive model.

## 3.3. Screening the best performing model

This stage entails selecting which ML model, among those trained, will be utilized for technical evaluation. Typically, the evaluation assesses how well the model identifies valuable patents in terms of accuracy, precision, recall, and F1-score. However, in the domain of technology valuation, where valuable patents are fewer in number compared to less valuable ones, the dataset for ML models tends to be imbalanced. Hence, additional metrics such as MCC or Youden's J may be considered for model selection. Moreover, to validate the reliability of the model's predictions, the ECE can be utilized. These metrics may carry varying weights depending on the context or purpose of model usage. To address this, we propose utilizing the Pareto-front map, offering a boundary for selecting the most suitable model. Details for specific performance measures are below.

ML-based technology valuation models aim to screen valuable patents from a vast pool, which often leads to a strong emphasis on metrics such as accuracy, precision, recall, and F1-score. As most technology valuation datasets are likely to be imbalanced, it is important to check the confusion matrix, where True represents valuable patents and False indicates non-valuable patents. In the confusion matrix, True Positive (TP) corresponds to correctly classified valuable patents. True Negative (TN) represents correctly classified non-valuable patents. False Negative (FN) does valuable patents that are incorrectly classified as non-valuable, which might result in missing out on potentially significant patents, impacting the model's ability to identify valuable innovations. False Positive (FP) includes non-valuable patents misclassified as valuable, leading to a false sense of significance or value. FP might lead to unnecessary attention or investment directed towards patents that aren't truly valuable, while FN could result in overlooking genuinely valuable patents, affecting decision-making processes regarding technological innovations and investments. Precision is measured based on the proportion of correctly identified valuable patents among all patents classified as valuable by the model, while recall is based on the ratio of correctly predicted positive observations to all actual positives. Accuracy is calculated as (TP + TN) / (TP + TN + FP + FN), representing the ratio of correctly classified patents (both valuable and non-valuable) to the total number of patents.

In datasets characterized by class imbalances, traditional performance metrics such as accuracy may not accurately reflect classifier efficacy. This is because classifiers can achieve seemingly high accuracy by favoring the majority class while neglecting the minority class. Therefore, alternative evaluation measures are crucial for a more nuanced understanding of classifier performance. Two widely used metrics in such scenarios are Youden's J statistic and the Matthews Correlation Coefficient (MCC). Youden's J statistic combines sensitivity (true positive rate) and specificity (true negative rate) into a single metric, offering a comprehensive assessment of classifier performance. It is expressed as:

$$J = \frac{TP}{TP + FN} + \frac{TN}{TN + FP} - 1 = sensitivity + specificity - 1$$

which ranges from 0 to 1, with higher values indicating better overall performance. Similarly, MCC is a robust metric that considers all elements of the confusion matrix, providing a balanced evaluation of classifier performance across all classes. It is calculated as:

$$MCC = \frac{TP \times TN - FP \times FN}{\sqrt{(TP+FP)(TP+FN)(TN+FP)(TN+FN)}}.$$

Here, TP denotes true positives, TN denotes true negatives, FP denotes false positives, and FN denotes false negatives. MCC ranges from -1 to 1, with 1 indicating perfect classification, 0 indicating random classification, and -1 indicating total disagreement between prediction and observation. This metric provides a comprehensive evaluation of classifier performance, considering both sensitivity and specificity while accounting for class imbalances.

Assessing the reliability of a ML model involves evaluating the confidence in its predictions. One crucial metric for this assessment is the Expected Calibration Error (ECE) score, which measures how well the model's predicted probabilities align with the actual correctness of those predictions.

$$ECE = \sum_{m=1}^{M} \frac{|B_m|}{n} |acc(B_m) - conf(B_m)|$$

In the context of ML-based technology valuation models, reliability is pivotal. Reliable models not only deliver accurate predictions but also provide well-calibrated and trustworthy confidence estimates. This reliability is crucial for critical decisions regarding technological advancements, investments, or research directions. In the realm of technology valuation, the model reliability ensures the credibility of insights derived, influencing strategic decisions and resource allocations in technological domains.

### 3.4. Interpreting the model's predictions.

After deriving prediction results for valuable patents from the selected best performing ML model, we use SHAP to interpret the results. Interpreting the results of a prediction model accurately is important because it establishes appropriate user trust and provides insight into how the model can be improved. Some previous studies have used simple models (e.g., linear models) rather than complex models because they are easier to interpret, even if they are less accurate. However, as the availability of big data has increased, the benefits of using complex models have also increased, and it is not only the accuracy of the prediction results that is important, but also the interpretability. To complement the interpretability of complex models, various interpretation methods have been proposed to explain how specific features contribute to the prediction results and the behavior of the model (Bach et al., 2015; Ribeiro et al., 2016; Štrumbelj and Kononenko, 2014). Among the multiple interpretation methods, SHAP is a popular unified framework for interpreting predictions of ML models based on the Shapley value of the conditional expectation of a model (Lundberg and Lee, 2017). SHAP is able to facilitate better understanding of ML model behavior by assigning each feature an importance value for a particular prediction.

Given the original predictive model $f$ to be explained and the explanation model $g$, SHAP measures the importance of features based on additive feature attribution methods:

$$g(z') = \phi_0 + \sum_{i=1}^{M} \phi_i z'_i$$

where $z' \in \{0, 1\}^M$ denotes the vector of simplified input features obtained from the original input features $x$; $M$ denotes the number of input features; $\phi_i \in R$ indicates the importance value of the $i$th feature; and $\phi_0$ is the model output without any feature. To be specific, SHAP estimates the importance of each feature as a change in the expected model prediction conditional on that feature and explains how to change from a baseline value $E[f(z)]$ to the current output $f(x)$. If the features are not independent, the order in which the features are added to the expected value is important, and SHAP averages the value of $\phi_i$ over all possible orders. So, when defining $f_x(S) = E[f(x)|x_S]$ for a subset of features ($S$), the SHAP value ($\phi_i$) is expressed as:

$$\phi_i = \sum_{S \subseteq \{x_1, \dots, x_m\} \setminus \{x_i\}} \frac{|S|!\,(M - |S| - 1)!}{M!} (f_x(S \cup \{x_i\}) - f_x(S))$$

where $f_x(S \cup \{x_i\})$ and $f_x(S)$ are the model prediction with and without the $i$th feature, respectively. Based on cooperative game theory, which uses Shapley values to measure attribution of each player to the game result, the attribution of the $i$th feature is calculated as the mean difference of $f_x(S \cup \{x_i\})$ and $f_x(S)$ for all possible subsets of simplified model features. Therefore, the SHAP values represent the importance, which is the contribution of each feature to the model prediction. In this study, SHAP presents the influence of patent indicators on the ML model's prediction of technology value.

# 4. Empirical study and results

## 4.1. Overview

We conducted a case study of semiconductor technology for several reasons. First, the value of patents in the semiconductor industry is linked to industrial significance and economic value, as individual patents impact various industries such as integrated circuits, memory technologies, sensors, power management, and field-programmable gate arrays. Second, considering the rapidly increasing number of patents in the semiconductor industry, it is crucial for companies to discern valuable technologies and selectively maintain patents in terms of mitigating infringement risks and economic viability. Finally, as the semiconductor industry grows both quantitatively and qualitatively, practitioners in this field call for a scientific data-driven approach to obtain objective information on valuable technologies. Therefore, it is necessary to analyze the extensive patent information on semiconductor technology to identify valuable patents and further promising technology areas.

## 4.2. Calibrated ML approach to technology valuation

### 4.2.1. Extraction of patent indicators and patent lifetime

For the case study, we searched patents of which titles, claims, or abstracts include keywords such as 'semiconductor' and IPC code includes 'H01L' ("semiconductor devices; electric solid-state devices not otherwise provided for") from the USPTO. Specifically, we collected bibliometric information, textual information as well as maintenance history, using data processing service provided by PatentsView of the USPTO. Consequently, 74,043 patents registered during 2000-2019 with the determined lifetimes were used for this experiment. To simplify the problem of this model, we aimed to classify whether the patent lifetime is maximum or not with patent indicators. Patents with lifetime of 4, and maximum years were used and their counts were 12,639, and 34,334, respectively. The results of this step are not reported here in its entirety owing to lack of space, but part of them is shown as Table 2.

**Table 2.** Parts of input matrix of non-valuable and valuable patents

| Patent number | TE_5 | SC_1 | PK_4 | … | CP_4 | SC_6 | PK_6 | Label |
|---|---|---|---|---|---|---|---|---|

| | | | | | | | |
|---|---|---|---|---|---|---|---|
| 6010919 | 1158 | 4442 | 17 | ... | 1006 | 84 | 0.321 | VP |
| 6013579 | 1140 | 840 | 315 | ... | 447 | 158 | 0.150 | VP |
| 6121821 | 875 | 3298 | 1643 | ... | 539 | 248.4 | 0.526 | VP |
| 6245676 | 579 | 5298 | 1921 | … | 841 | 246.8 | 0.733 | VP |
| 7083994 | 2247 | 2017 | 0 | … | 1394 | 254 | 0.534 | VP |
| 6037626 | 1176 | 1465 | 524 | … | 606 | 92 | 0.594 | NVP |
| 6137149 | 989 | 1790 | 1680 | … | 1215 | 151.7 | 0.489 | NVP |
| 6503771 | 5909 | 3602 | 777 | … | 1169 | 93 | 0.505 | NVP |
| 6898544 | 6707.5 | 11291 | 1811 | … | 413 | 221 | 0.373 | NVP |
| 7309901 | 3821 | 1999 | 2947 | … | 965 | 285 | 0.511 | NVP |

**Notes:** VP denotes valuable patents, while NVP represents non-valuable patents.

### 4.2.2. Valuation of technological value of patents

We developed the five ML models using pytorch, XGBoost, H2O AutoML packages in python language, and their hyperparameters were carefully determined to obtain the best performance of each ML model. Here, we aimed to determine the hyperparameters using 10-fold cross-validation, utilizing the F1-score as an overall evaluation metric. For example, in developing the NN model, we adjusted the number of hidden layers, the number of nodes, and the learning rate to optimize performance. The final model architecture consists of one hidden layer, which contains 100 nodes with rectified linear unit function. We used an Adam optimizer with a learning rate of 0.005. Dropout was introduced before the hidden layer. The LR model was trained for 100 epochs using the solver 'IRLSM' with the Elastic-Net combining Lasso and Ridge regularization. The RF model utilized 50 trees with the maximum depth of 20, while the XGB model utilized 90 estimators, with a maximum exploration depth of 6 and a learning rate of 0.3.

Regarding input data configuration, to alleviate the burden of imbalanced datasets and potential noises, we adopted a kind of under-sampling method, Tomek Links, which aims to eliminate overlapping regions at the border between these classes. This process helps in reducing noise around the minority class, potentially enhancing the model's ability to discern class boundaries. In mathematical terms, suppose $X$ represents the dataset, $x_i$, denotes individual data points, and $y_i$ represents the corresponding class labels. The goal is to find the pairs $(x_a, x_b)$ such that $x_a$ belongs to the minority class $(y_a)$, $x_b$ belongs to the majority class $(y_b)$, and $x_a$ is the nearest neighbor of $x_b$. This process aids in enhancing the model's ability to learn and generalize effectively, especially in scenarios with imbalanced label data, ultimately contributing to improved performance in technology valuation assessment models.

Parts of the evaluation results are given owing to lack of space in Table 3. The last four columns represents the prediction results of each ML model, indicating that the technological values of individual patents were sometimes differently judged by model.

**Table 3.** Part of the evaluation results of ML models.

| Patent number | Input | | | | Label | Prediction | | | |
|---|---|---|---|---|---|---|---|---|---|
| | TE_5 | SC_1 | PK_4 | ... | | LR | RF | NN | XGB |
| 6010919 | 1158 | 4442 | 17 | … | VP | VP | VP | VP | VP |
| 6013579 | 1140 | 840 | 315 | … | VP | VP | VP | VP | NVP |
| 6121821 | 875 | 3298 | 1643 | … | VP | VP | VP | NVP | NVP |
| 6245676 | 579 | 5298 | 1921 | | VP | VP | VP | VP | NVP |
| 7083994 | 2247 | 2017 | 0 | … | VP | VP | VP | VP | VP |
| 6037626 | 1176 | 1465 | 524 | … | NVP | VP | NVP | VP | NVP |
| 6137149 | 989 | 1790 | 1680 | … | NVP | VP | NVP | NVP | NVP |
| 6503771 | 5909 | 3602 | 777 | … | NVP | NVP | NVP | NVP | NVP |
| 6898544 | 6708 | 11291 | 1811 | … | NVP | NVP | NVP | NVP | NVP |
| 7309901 | 3821 | 1999 | 2947 | … | NVP | VP | NVP | NVP | NVP |

### 4.2.3. Evaluation of model performance

The ML models with each optimal hyperparameter, were obtained through 10-fold cross-validations. We evaluated their performance with basic measures such as accuracy, precison, recall and F1-score. Considering that the dataset is imbalanced, we employed Youden's $J$ statistics and MCC. In Table 4, the performance of each optimized ML model is presented and others are reported in Table A.1.

Our analysis reveals that the majority of ML models achieve an impressive accuracy of 90-91% when tasked with distinguishing between valuable technologies and non-valuable technologies. Upon careful evaluation of precision, recall, and F1 score, the XGB model emerged as the marginal frontrunner in performance. Nevertheless, in light of the dataset's inherent imbalance with fewer instances of valuable technologies compared to non-valuable ones, we further scrutinized our findings by incorporating J statistics and MCC. Consequently, XGB and RF models were identified as the top-performing models based on $J$ statistics and MCC, respectively. This suggests that XGB is particularly effective in balancing sensitivity and specificity, while RF excels in managing imbalanced datasets and delivering robust binary classifications. These findings underscore the strengths of each model in different evaluation metrics, highlighting their suitability for various classification tasks.

**Table 4.** Model performance comparison of the best version of ML models.

| Model | Accuracy | Precision | Recall | F1-score | Youden's J | MCC |
|---|---|---|---|---|---|---|
| LR | 0.897 | 0.883 | 0.991 | 0.934 | 0.634 | 0.732 |
| RF | 0.905 | 0.906 | 0.971 | 0.937 | 0.697 | 0.750 |
| NN | 0.901 | 0.892 | 0.984 | 0.936 | 0.660 | 0.741 |
| XGB | 0.908 | 0.904 | 0.978 | 0.939 | 0.695 | 0.758 |

To assess the prediction reliability of each ML model, we utilized reliability diagrams. These diagrams provide a visual representation of model performance based on predicted probabilities (Fig. 2). The horizontal axis represents the predicted probabilities, while the vertical axis indicates the fraction of positives. Due to the 10-fold cross-validation method, ten curves are depicted for each model. The thick black line represents the mean of these ten curves, and the shaded gray area denotes the standard deviation. As illustrated in the figure, the LR model deviates significantly from the diagonal line, tending to overconfident in the range below 50% and slightly underconfident elsewhere. The reliability curve of the RF model shows that the calculated percentages slightly overestimate the rates of the test data in the 20-60% range, but align closely with the diagonal line outside this range. In contrast, the NN model tends to underestimate in the range below 50%, but matches the diagonal line thereafter. And the XGB model demonstrated a consistent alignment with the diagonal across almost the entire range, with the exception of the 40-60% range. Their reliabilities were also measured based on ECE scores. The LR model achieved a ECE of 0.175, while the RF, NN and XGB models showed 0.188, 0.197 and 0.203, respectively.

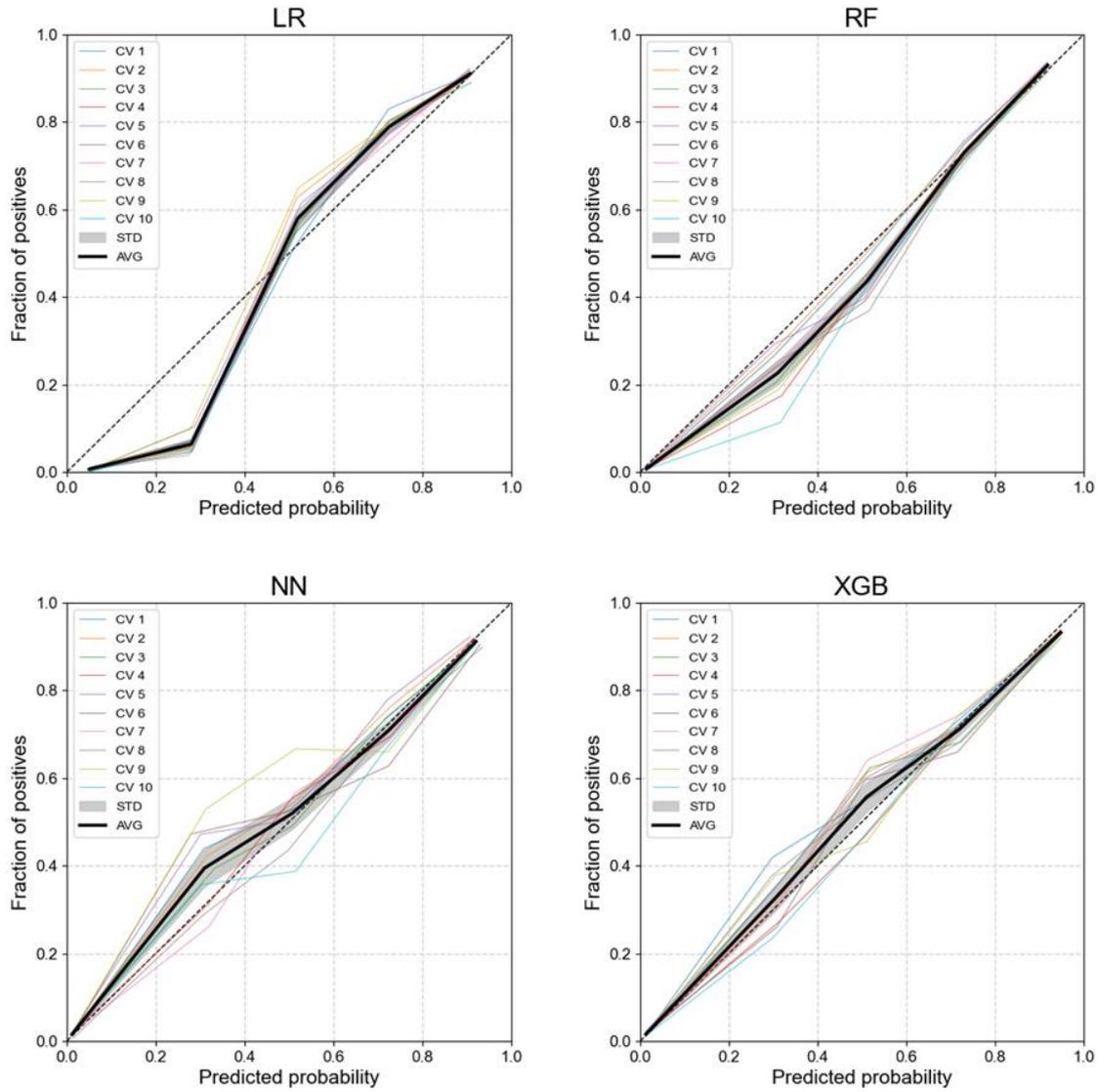

**Fig. 2.** Reliability diagrams of ML-based models

We observed that the majority of the developed ML models achieved a high F1-score of approximately 93%. In addition to this, we evaluated the performance metric (i.e., MCC) and reliability (i.e., ECE) while considering the data distribution. Selecting the appropriate ML model can vary based on the analysis environment or specific objectives; therefore, we present the Pareto-front map of the ML models (Fig. 3). During the parameter search process, we assessed models that demonstrated performance above a certain threshold from the perspectives of MCC and ECE. An ECE score closer to zero and an MCC score closer to one signify superior models. We delineated the explored models' range as feasible points. Furthermore, we identified the optimal model selection frontier by marking the Pareto-front and subsequently selected the optimized RF model for further application in the next stage.

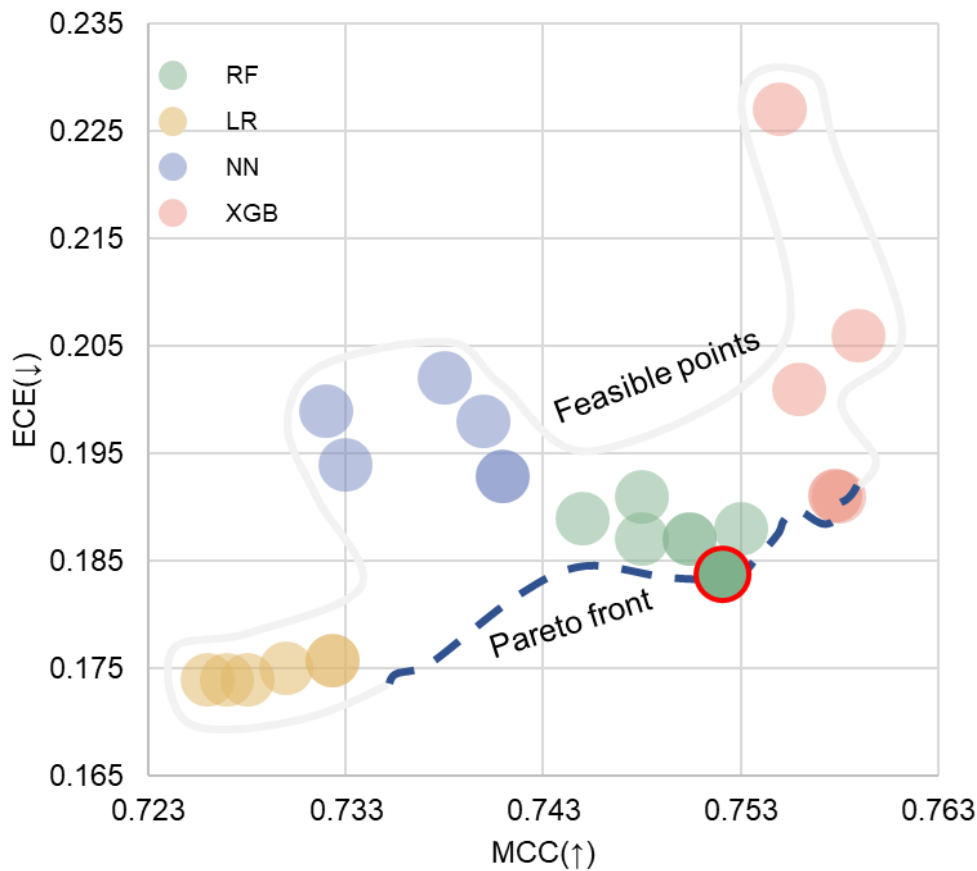

**Fig. 3.** Pareto-front map of ML-based technology valuation models.

**Notes:** Each node's color represents a different model, with the darker-colored nodes indicating the models that achieved the highest F1-score within their respective categories. The final selected model is highlighted with a red border.

### 4.2.4. Interpretation of model predictions

Fig. 4 presents the summary plot and the box plot for confidence intervals of the best-performing model (i.e., RF). The analysis identified TE_1, TE_3, TE_2, PK_2, and PK_5 as significant factors influencing the likelihood of a patent being maintained for an extended period. In Fig. 4, the regions where SHAP values are negative contain numerous red dots, signifying that the values of these features exhibit a negative correlation with their corresponding SHAP values. This indicates that in mature technology fields characterized by large scale, intense competition, and high activity levels, patents are less likely to be maintained for extended periods if the applicant has a significant history of patenting outside the semiconductor domain.

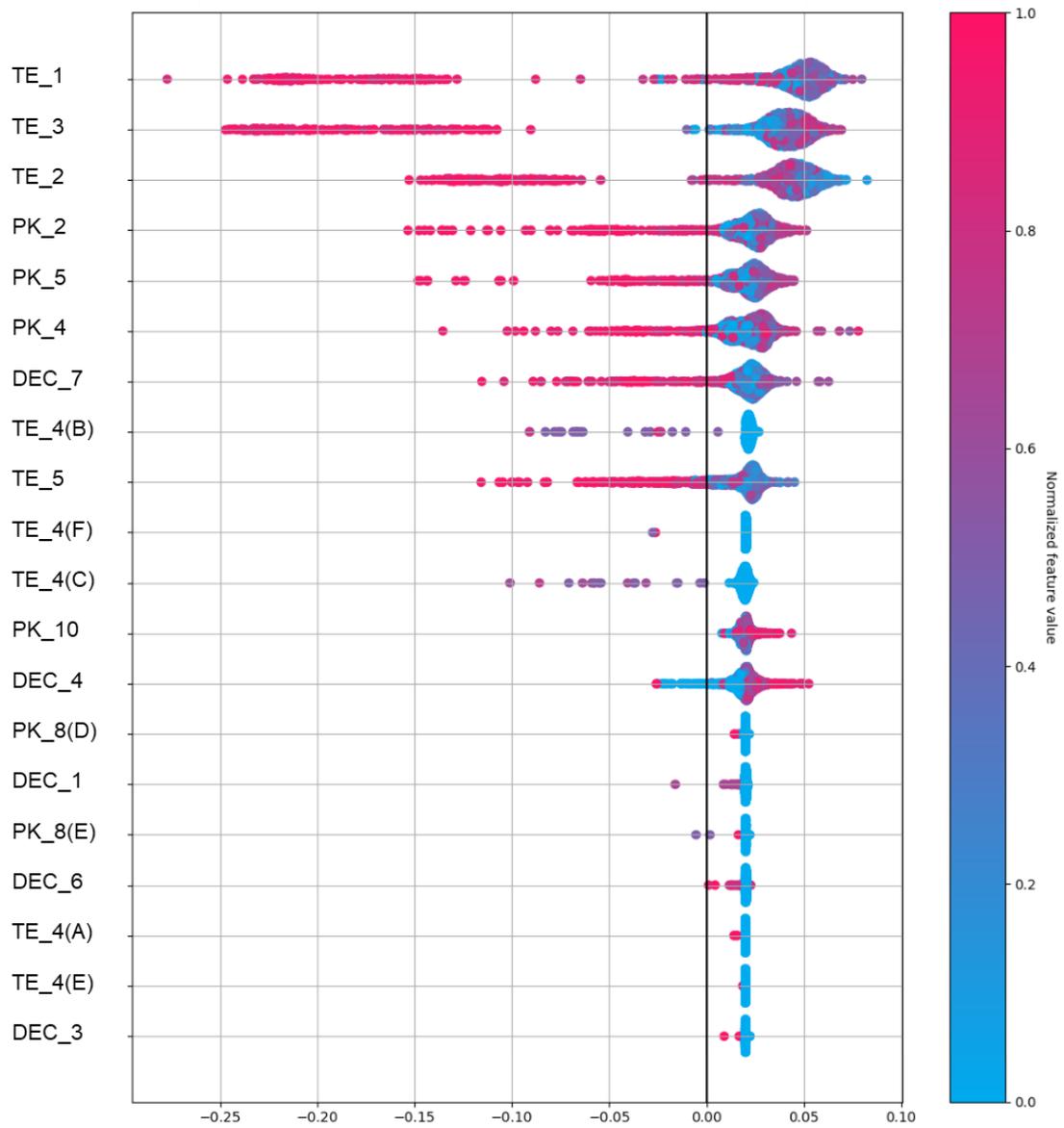

**Fig. 4.** Results of SHAP analysis such as summary plot at the global level, where the horizontal axis is SHAP value and the vertical axis is the significant input feature.

Further, we identified the key features for each confidence bin, as shown in Fig. 5a-b. In the low confidence bin, features such as PK_8(E), TE_4(A), TE_4(E), PK_8(A) and PK_8(B) exhibit positive relationships with their SHAP values (Fig. 5a). In uncertain areas, the extent to which current or prior knowledge is related to specific technology fields influences the probabilities of long-term maintenance. Conversely, features such as TE_2, TE_1, and TE_3 have negative relationships with their SHAP values (Fig. 5b), indicating that larger size, greater activity, and heightened competitiveness of relevant technologies decrease the probability of the technology being highly valuable. In confident areas with high confidence of 0.8 ~ 1.0, features such as TE_1, TE_2, TE_3, PK_2 and PK_5 are observed as main features with strong positive relationships with their SHAP values. For each confidence bin, the ranking of input features with high impact on SHAP values is identified and more details can be observed in Table A.2.

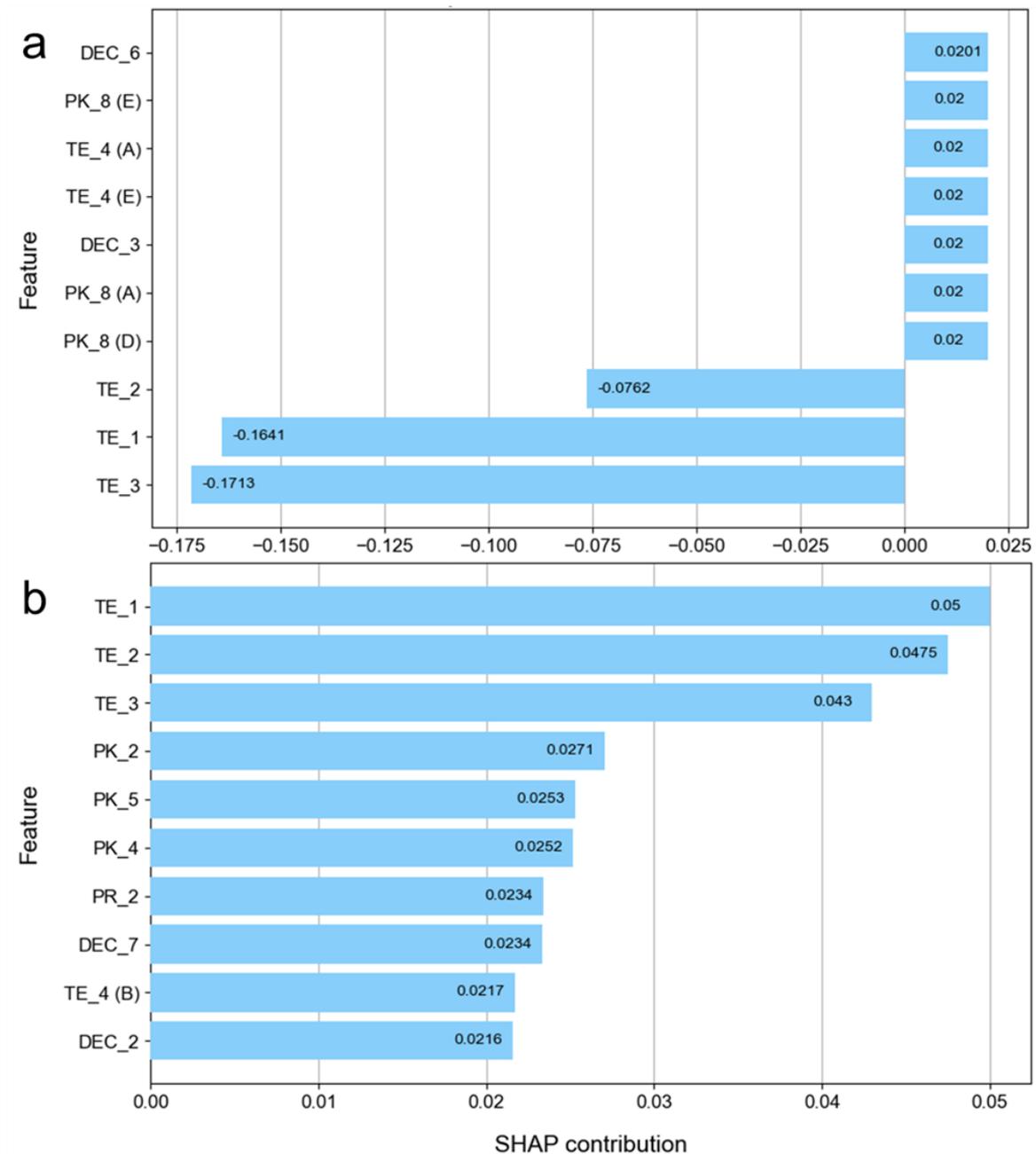

**Fig. 5.** Results of SHAP analysis such as bar plots of low confidence bin (0.0~0.2), and of high confidence (0.8~1.0).

# 5. Discussion

## 5.1. Implications for theory and practice

The proposed approach offers new insights into developing reliable and accurate ML models for technology valuation, with significant theoretical and practical implications. First, it enhances previous ML methods by creating an interactive framework that allows experts to assess and interpret technology valuation results based on their knowledge and judgment, contributing to the literature on technology valuation. Unlike earlier models that focused solely on distinguishing valuable technologies from less valuable ones, this approach provides a robust quantitative estimation of technology value, improving our understanding of uncertainty. To achieve this, we established a framework capable of assessing the uncertainty of technology valuation results and interpreting their underlying causes through quantitative evaluations of confidence in trained ML models. To the best of our knowledge, this study is a pioneering attempt to analyze the reliability of ML-based technology valuation models quantitatively. Specifically, SHAP analysis by confidence interval segments has enhanced our understanding of uncertainty by revealing differences in important indicators between high and low certainty areas. The proposed method facilitates successful collaboration between experts and machines by conducting technology valuation with interpretable and reliable quantitative results. Additionally, the systematic process and scientific methods used in this study pave the way for developing automated platforms for technology valuation. While the primary focus was on determining which patents would endure over their maximum period, the proposed method and findings are versatile and can also serve as valuable tools for assessing technological impact and economic value, broadening their applicability and significance across various domains.

Second, the proposed approach has significant practical implications, as its systematic process can be reproduced and adapted across various technology domains. This supports the technology valuation process for domain experts who may lack ML model knowledge or skills. The proposed approach can be developed into an interactive software system that generates intermediate results at each stage: selecting patent indicators, conducting technology valuations, and determining the best-performing model based on expert feedback. While the developed software can provide evaluation results for user input datasets using pre-trained ML models, achieving superior performance by adapting the model to specific domains requires considerable time and effort. Industrial practitioners need to periodically update their training datasets to incorporate relevant patent indicators and reflect the latest technological trends in their specific domains or for particular analytical objectives. The proposed approach facilitates this by allowing users to adjust input features and develop customized technology valuation models, offering reliable assessments and interpretative results. Once a sufficient dataset for the area of interest is secured, the best-performing model can be identified and interpreted following the proposed procedure. This ensures that the model remains relevant and effective, providing accurate and actionable insights.

## 5.2. Implementation and customization of the proposed approach

This study proposes a systematic approach to developing reliable and accurate ML models for technology valuation, utilizing quantitative outcomes and scientific methods. The proposed approach offers several advantages over previous studies: First, the proposed approach enables experts to collaborate effectively with ML models by providing quantitative and interpretable outcomes on prediction uncertainty, even if there is a discrepancy between model predictions and expert judgments. Second, the proposed approach serves as a valuable complementary tool to support experts' decision-making in the domain-specific technology valuation process, addressing the highly uncertain and vulnerable technological environment. The interactive steps, which allow experts to consider more technological characteristics, define input features, and crystallize technology values based on their knowledge and judgment, are embedded throughout the approach. This enhances its practical implications compared to previous methods. Third, the outcomes offer more practical assistance with decision-making in the technology valuation process than previous methods. Unlike a binary classification of valuable or not, the value of technology is provided as a reliable number between 0 and 1, offering a more nuanced and actionable assessment.

However, the newly developed method should be deployed with caution in practice. Industrial practitioners should consider several critical factors when applying the proposed approach. The following considerations are paramount: (1) industrial practitioners should examine as many input features as possible within their accessible technology landscape to ensure the successful deployment of ML models, and (2) the implications of technology value identified in this study differ significantly from those addressed in prior works. The outcomes of the proposed approach can broaden experts' horizons and enrich their understanding of evaluating valuable technologies. The proposed and existing approaches are not mutually exclusive and can be combined. For instance, in defining patent indicators and proxies of technology value, prior approaches such as using text as input and forward citations as output can help crystallize valuable technologies. The technology valuation process presented in this study is designed as an interactive and adaptable structure for predicting the future value of technologies. For implementation and customization in a given domain, it should be systematic, generating a set of input features that comprehensively reflect the nature of the technology domain. Patent indicators represent the scope of exploring technological characteristics, and thus the outcomes, including performance, may differ depending on these setting. Although the proposed method effectively supports the labor-intensive and time-consuming technology valuation process, it still requires some expert intervention and manual work when interpreting the identified feature importance to validate the models' predictions. The value of the results can be enhanced when integrated with other technology valuation methods and evaluation metrics. For example, domain-specific measures such as the Jonckheere–Terpstra tests and data envelopment analysis can be used to explore different implications of quantitative indicators for identified valuable technologies (Kim et al., 2023). The proposed approach includes alternative methods that can aid in understanding valuable technologies, such as SHAP values of input features and the ECE score of predictions. While the use of SHAP and ECE is appropriate given the size of the training dataset and the purpose of this study, improvements in model interpretation and uncertainty quantification will be necessary in future works (Lee and Kim, 2022).

## 6. Conclusion

This study presents an analytical framework for developing reliable technology valuation models using calibrated ML techniques. The core premise of this research is that the reliability of ML models is as important as their high accuracy, particularly for enhancing practical value in high-uncertainty applications. To achieve this, we aim to identify well-calibrated ML models through ECE score analysis and examine the impact of input features on uncertainty using SHAP analysis. The validity of our proposed approach is illustrated through a case study on semiconductor technology patents registered with the USPTO.

The contributions of this study are two-fold. Firstly, from an academic standpoint, it represents a pioneering effort to scrutinize the reliability of ML-based technology valuation models amid their increasing adoption, thereby shifting the focus from mere accuracy to the crucial aspect of reliability. Additionally, the integrated use of ECE score analysis and SHAP analysis constitutes a methodological breakthrough, extending applicability beyond technology valuation to various fields requiring dependable evaluation models. Secondly, from a practical standpoint, our systematic framework shows promise in aiding technology experts across diverse domains to seamlessly assess their innovations and identify suitable models. By leveraging patent lifetime as a universal proxy for technology value, our approach offers a more domain-agnostic alternative compared to conventional proxies such as forward citations. This not only enhances the robustness of ML-based technology valuation models but also paves the way for their broader applicability and adoption.

Despite its significant contributions, this study is subject to several limitations that should be addressed in future research. Firstly, it does not address case studies specific to individual fields, limiting the understanding of the relationships between reliability and domain specificity. Expanding case studies presents a valuable research opportunity, as key models, indicators, and value assessment proxies may vary significantly across different domains. Secondly, the study does not explore differences in reliability arising from variations in model parameters or techniques. While reliability from a model architecture perspective may not be a primary concern in the technology valuation field compared to computer science (Minderer et al., 2021), it remains a worthwhile subject for exploration, particularly regarding the impact of data sampling methods on reliability. Additionally, the study does not investigate the reliability of various proxies for technology values. Although domain-specific proxies such as citations and technology transfers were excluded due to their varying distributions, examining the reliability of these indicators could be valuable for specific fields where they are prevalent. Finally, expanding the quantitative range of reliability indicators, such as negative log loss and brier score, represents another important area for further research.

## Acknowledgements

This work was supported by the National Research Foundation of Korea (NRF) funded by the Ministry of Education (NRF-2021R1A2C1010027) and the Human Resources Program in Energy Technology of the Korea Institute of Energy Technology Evaluation and Planning

# Appendix

A1. ML model descriptions

First, LR is a supervised ML model that, unlike linear regression, performs binary classification tasks by predicting the probability of an outcome between 0 and 1. LR is applied to evaluate the relationship between a dependent binary variable with one or more independent variables at the nominal, ordinal, interval, or ratio level (Peng et al., 2002). Second, RF is an ensemble ML model consisting of a combination of various decision trees. Each tree that constitutes RF casts a unit vote for the most popular class at the input, and the final predictive value of RF is estimated by considering the majority voting system (Breiman, 2001). RF is utilized in a wide variety of domains due to its ability to perform multi tasks such as regression and supervised/unsupervised classification (Biau and Scornet, 2016). Third, NN is a ML model which transforms a set of input features into a set of output classes, and is particularly effective for datasets with non-linear relationships. NN is a foundational model in deep learning, inspired by the structure and function of the human brain (Hinton and Salakhutdinov, 2006). A typical neural network consists of an input layer, one or more hidden layers, and an output layer (Liu et al., 2017). Each layer contains a number of neurons (nodes) that are fully connected to the neurons in the previous and next layers. Backpropagation is the key learning mechanism in neural networks, which involves computing the gradient of the loss function with respect to each weight and bias in the network by applying the chain rule of calculus, effectively determining how the loss would change if the weights and biases are adjusted. Lastly, XGB is an optimized gradient tree boosting model to improve the learning process (Chen and Guestrin, 2016). In this approach, it constructs a new model that predicts the gradients (or residuals) of the loss function, regarding the predictions made by the existing sequence of models. A key feature of XGB is the inclusion of regularization terms (both L1 and L2), which helps to control overfitting, making it superior to standard gradient boosting which might not include regularization.

**Table A1.** Model performance comparison of all ML models.

| Model | Accuracy | Precision | Recall | F1-score | Youden's J | MCC | ECE |
|---|---|---|---|---|---|---|---|
| RF #1 | 0.904 | 0.904 | 0.973 | 0.937 | 0.691 | 0.748 | 0.187 |
| RF #2 | 0.906 | 0.905 | 0.973 | 0.938 | 0.697 | 0.753 | 0.188 |
| RF #3 | 0.903 | 0.901 | 0.975 | 0.936 | 0.683 | 0.745 | 0.189 |
| RF #4 | 0.904 | 0.902 | 0.975 | 0.937 | 0.686 | 0.748 | 0.191 |
| LR #1 | 0.896 | 0.883 | 0.989 | 0.933 | 0.632 | 0.727 | 0.174 |
| LR #2 | 0.895 | 0.882 | 0.989 | 0.932 | 0.63 | 0.726 | 0.174 |
| LR #3 | 0.896 | 0.883 | 0.989 | 0.933 | 0.632 | 0.728 | 0.174 |
| LR #4 | 0.897 | 0.883 | 0.99 | 0.933 | 0.634 | 0.73 | 0.175 |
| NN #1 | 0.901 | 0.891 | 0.984 | 0.935 | 0.659 | 0.74 | 0.198 |
| NN #2 | 0.899 | 0.892 | 0.98 | 0.934 | 0.659 | 0.733 | 0.194 |
| NN #3 | 0.898 | 0.888 | 0.985 | 0.934 | 0.646 | 0.732 | 0.199 |
| NN #4 | 0.900 | 0.894 | 0.98 | 0.935 | 0.664 | 0.738 | 0.202 |
| XGB #1 | 0.908 | 0.904 | 0.978 | 0.939 | 0.695 | 0.758 | 0.191 |
| XGB #2 | 0.908 | 0.905 | 0.977 | 0.939 | 0.698 | 0.759 | 0.206 |
| XGB #3 | 0.907 | 0.905 | 0.975 | 0.939 | 0.697 | 0.756 | 0.201 |
| XGB #4 | 0.907 | 0.905 | 0.974 | 0.939 | 0.698 | 0.755 | 0.227 |

**Notes**: RF #1 and #2 were trained with 50 trees (max depths of 20 and 15), RF #3 with 40 trees (max depth of 10), and RF #4 with 20 trees (max depth of 10). LR #1 and LR #2 used Ridge regularization ($\lambda$=0.0081) for 36 epochs, while LR #3 and #4 utilized Elastic-Net regularization ($\alpha$=0.5) for 32 and 33 epochs respectively with $\lambda$ values of 0.0062 and 0.0047. NN #1 had 100 nodes and 10% dropout in one hidden layer, NN #2 had the same configuration without dropout, NN #3 had 50 nodes with 40% dropout, and NN #4 had 100 nodes with 40% dropout. XGB #1 to XGB #4 were configured with 75, 61, 61, and 54 estimators, respectively.

**Table A.2.** Comparison of significant features for each confidence bin.

| | Confidence bin | | | | |
|---|---|---|---|---|---|
| Rank | 1st bin (0.0 ~ 0.2) | 2nd bin (0.2 ~ 0.4) | 3rd bin (0.4 ~ 0.6) | 4th bin (0.6 ~ 0.8) | 5th bin (0.8 ~ 1.0) |
| 1 | TE_3 | PK_2 | PK_5 | TE_2 | TE_1 |
| 2 | TE_1 | PK_5 | PK_2 | TE_1 | TE_2 |
| 3 | TE_2 | DEC_7 | TE_2 | TE_3 | TE_3 |
| 4 | DEC_6 | SC_5 | TE_1 | PK_8(D) | PK_2 |
| 5 | PK_8(E) | DEC_6 | PK_6 | TE_4(E) | PK_5 |
| 6 | TE_4(A) | PK_8(A) | PK_8(D) | DEC_6 | PK_4 |
| 7 | TE_4(E) | DEC_1 | DEC_6 | DEC_3 | PR_2 |
| 8 | DEC_3 | TE_4(E) | TE_4(A) | TE_4(A) | DEC_7 |
| 9 | PK_8(A) | DEC_3 | TE_4(E) | PK_8(E) | TE_4(B) |
| 10 | PK_8(D) | PK_8(D) | DEC_3 | DEC_1 | DEC_2 |